%% file: EMBS_2021.tex
\documentclass[letterpaper, 10 pt, conference]{ieeeconf}  

\IEEEoverridecommandlockouts                              

\usepackage{cite}

\usepackage[pdftex]{graphicx}
\graphicspath{{Figures/}{Figures/ExampleImages/}{Figures/CNNs/}}
\DeclareGraphicsExtensions{.pdf,.jpeg}
\usepackage{amsmath, amsfonts}
\usepackage{algorithmic}

\usepackage[caption=false,font=footnotesize]{subfig}

\usepackage{url}

\newcommand{\includeCroppedPdf}[2][]{%
    \IfFileExists{./#2-crop.pdf}{}{%
        \immediate\write18{pdfcrop #2 #2-crop.pdf}}%
    \includegraphics[#1]{#2-crop.pdf}}

\title{\LARGE \bf
Combining Image Features and Patient Metadata to Enhance Transfer Learning
}

\author{Spencer A. Thomas$^{1}$ 
\thanks{$^{1}$Spencer A. Thomas is with the Data Science group, National Physical Laboratory, Teddington, UK
{\tt\small spencer.thomas@npl.co.uk}
}%
}

\begin{document}

\maketitle
\thispagestyle{empty}
\pagestyle{empty}

\begin{abstract}
In this work, we compare the performance of six state-of-the-art deep neural networks in classification tasks when using only image features, to when these are combined with patient metadata. We utilise transfer learning from networks pretrained on ImageNet to extract image features from the ISIC HAM10000 dataset prior to classification. Using several classification performance metrics, we evaluate the effects of including metadata with the image features. Furthermore, we repeat our experiments with data augmentation. Our results show an overall enhancement in performance of each network as assessed by all metrics, only noting degradation in a vgg16 architecture. Our results indicate that this performance enhancement may be a general property of deep networks and should be explored in other areas. Moreover, these improvements come at a negligible additional cost in computation time, and therefore are a practical method for other applications.
\end{abstract}




\input{intro_short}
\input{method_short}

\input{results_short}
\input{conclusion}

\section*{Acknowledgment}
The author would like to thank Nadia Smith and Peter Harris (NPL) for valuable feedback on this work. This work was funded by the Department of Business,
Engineering and Industrial Strategy through the cross-theme
national measurement strategy (Digital Health, 121572).

\bibliographystyle{IEEEtran}
\bibliography{refs}

\end{document}

%% file: intro_short.tex
\section{Introduction}
Deep learning has emerged as a powerful suite of tools for image classification \cite{LeCun2015}, and has a huge potential to solve challenges in healthcare settings. The use of deep neural networks is successful at tasks such as classification of medical images \cite{litjens_survey_2017}, analysis of electronic health records \cite{Thomas2019CMR, deLusignan2019, Avati2018} and segmenting data from emerging medical technologies \cite{Thomas2017, Behrmann2018}. This enormous potential comes with the caveat that very large amounts of data are required to train robust models that generalise beyond the training set. This requirement is unfortunately difficult to satisfy in the majority of biological and medical studies due to barriers to data availability.


Transfer learning has emerged as a promising method for circumventing the need for vast amounts of data to train deep networks \cite{Lakhani2018}. For domains with limited data, transfer learning utilises networks pre-trained on similar tasks with large amounts of data \cite{Pan2010}. Transfer learning is often used in medical imaging \cite{Xu2017, litjens_survey_2017, Shin2016} due to the limited availability of data that require expert labeling \cite{Rai2019transfer}. Transferring the image features from one domain to another can at least match the performance of models trained directly on the new domain \cite{tajbakhsh_convolutional_2016}. 
However the configuration of the transfer can be performed in a number of ways \cite{Rai2019transfer, Rai2019aggreated} and more research is needed in this area. 

Medical imaging data often has associated metadata used by clinicians in patient assessments. These metadata are multi type (numeric, categorical, etc) and are essential for maintaining the value of archived data \cite{Smith2019}. The information may be content related, e.g. scanner parameters, or relevant extracts from computerised medical records (CMR). These resources contain rich information relating to diseases \cite{de_lusignan_rcgp_2017,correa_royal_2016}, and data driven methods can identify patterns of patients \cite{Thomas2019CMR,deLusignan2019}. 

Classification tasks based on the combination of imaging with genomics data has been shown to surpass clinical experts in digital pathology \cite{Mobadersany2018}. 
Combining relevant information about the sample, e.g. patient demographics, with imaging data has also yielded high accuracy scores in binary classification tasks \cite{rocheteau_deep_2020}. However, the effect of combining these data is unknown and an assessment of any improvements or degradation to the networks in these frameworks is needed.

Clinicians will typically base diagnosis on several information sources either implicitly or explicitly. Demographic factors such as age can influence the likelihood of disease prevalence. In this work we investigate the combination of imaging data with related metadata to enhance classification performance evaluated by several metrics. We utilise transfer learning due to the limited volumes of data available, comparing the performance with and without metadata. Additionally we repeat the experiments with and without data augmentation during the training of the model. 

%% file: method_short.tex
\section{Methods}

A large collection of digital skin images from the International Skin Imaging Collaboration (ISIC) Melanoma Project \cite{ISIC} have been collated, processed and classified by expert dermatologists. The HAM1000 dataset from the ISIC database contains 10,015 digital images of skin lesions, each belonging to one of eight classes of skin conditions. Additionally the images have associated metadata containing clinical and acquisition information. The clinical fields contain a small amount of patient information including diagnosis of the images, an example is shown in Table~\ref{tab:metadata}.
Specifically, these are, age (numerical), sex (categorical) and anatomical site of the lesion (text). 

\begin{table}[h]
\renewcommand{\arraystretch}{1.3}
\caption{Image Metadata with ISIC Images}
\label{tab:metadata}
\centering
\begin{tabular}{|c||c|}
\hline
Clinical Field & Example Entry\\
\hline 
age approx & 55 \\
sex & female \\
anatom site general & lower extremity \\
\hline
melanocytic & true \\
benign malignant & malignant \\
diagnosis & melanoma \\
diagnosis confirm type & histopathology \\
 \hline 
\end{tabular}
\end{table}

\subsection{Deep Image Features} 

Consider the input data as $X \in \mathbb{R}^{N \times D}$ where $X_i$ is a $D$ dimensional data point with $N$ instances of the data. For imaging analysis $X$ is the imaging data with $D$ pixels and $N$ images. Deep learning takes $X$ as an input and applies a series of transformations through hidden layers typically in the form of convolutions. 
Following the notation of \cite{Vidal2017}, a matrix $W^k \in \mathbb{R}^{d_{k-1} \times d_k}$ is used to linearly transform the output of the $(k-1)$th layer, $X_{k-1} \in \mathbb{R}^{N \times d_{k-1}}$, into a $d_k$ -dimensional space, $X_{k-1}W^k\in \mathbb{R}^{N \times d_{k}}$, at the $k$th layer. 
The linear transformations are followed by a non-linear function, $\sigma_k(z)$, 
at each layer. The output of a network with $K$ layers is given by
\begin{equation}
\label{eq:deepLearning}
\mathcal{F} \left( X \right) =
\sigma_K \left( \dots 
\sigma_2 \left( \sigma_1 \left( X W^1 \right) W^2 \right) \dots W^K \right).
\end{equation}
$\mathcal{F} \left( X \right) \in \mathbb{R}^{N \times d_K}$, where $d_K$ is the dimensionality of $\mathcal{F} \left( X \right)$. For each network we select $K$ such that $\mathcal{F} \left( X \right)$ corresponds to the deepest set of image features, typically with the lowest dimensionality.

We compare several state-of-the-art deep convolutional neural network architectures for obtaining $\mathcal{F}$. All the networks used here have been pretrained using the ImageNet \cite{ImageNet} dataset, and  the network weights transferred to the ISIC image dataset. In this configuration, we are using the networks as feature extractors.  
Specifically we evaluate {alexnet} \cite{alexnet}, {densenet201} \cite{Huang2016}, {resnet50} \cite{He2015}, {inceptionresnetv2} \cite{Szegedy2017}, {vgg16} \cite{Simonyan2015} and googlenet \cite{googlenet} each with and without augmentation added to the input images. To account for the difference in input size to each network, all images are resized to the required dimensions using bi-linear interpolation. 

For the augmentation experiments, we introduce a subset of image manipulations, 
$X' = \Omega \left( X \right), $
where $\Omega$ represents the augmentation to the image prior to passing it to the network. The augmentation function introduces a random shift in the image of up to 30 pixels from its origin along the X axis and separately along the Y axis, random reflections in X and/or Y, and random rotations up to 90 degrees. This transformation is applied to the training and testing data.

\subsection{Integrating Images and Metadata}
The metadata for the images, $M$ are mapped such that they contain only numerical values to be compatible with standard neural networks. The mapping function $\mathcal{G} \left(M \right)$ converts the data to ASCII decimal introduced in \cite{Thomas2019CMR}. The conversion is performed element wise for an input string to allow maximum flexibility, for example, distinguishing upper and lower case letters, and mixed numerical and text inputs. When the input data differ in length, all instances are padded with trailing white space to the same size as the largest input string prior to conversion. Any missing entries in the fields are recorded as not a number (zero in ASCII decimal).

\begin{figure}[t]
\centering
\includegraphics[width=0.45\textwidth]{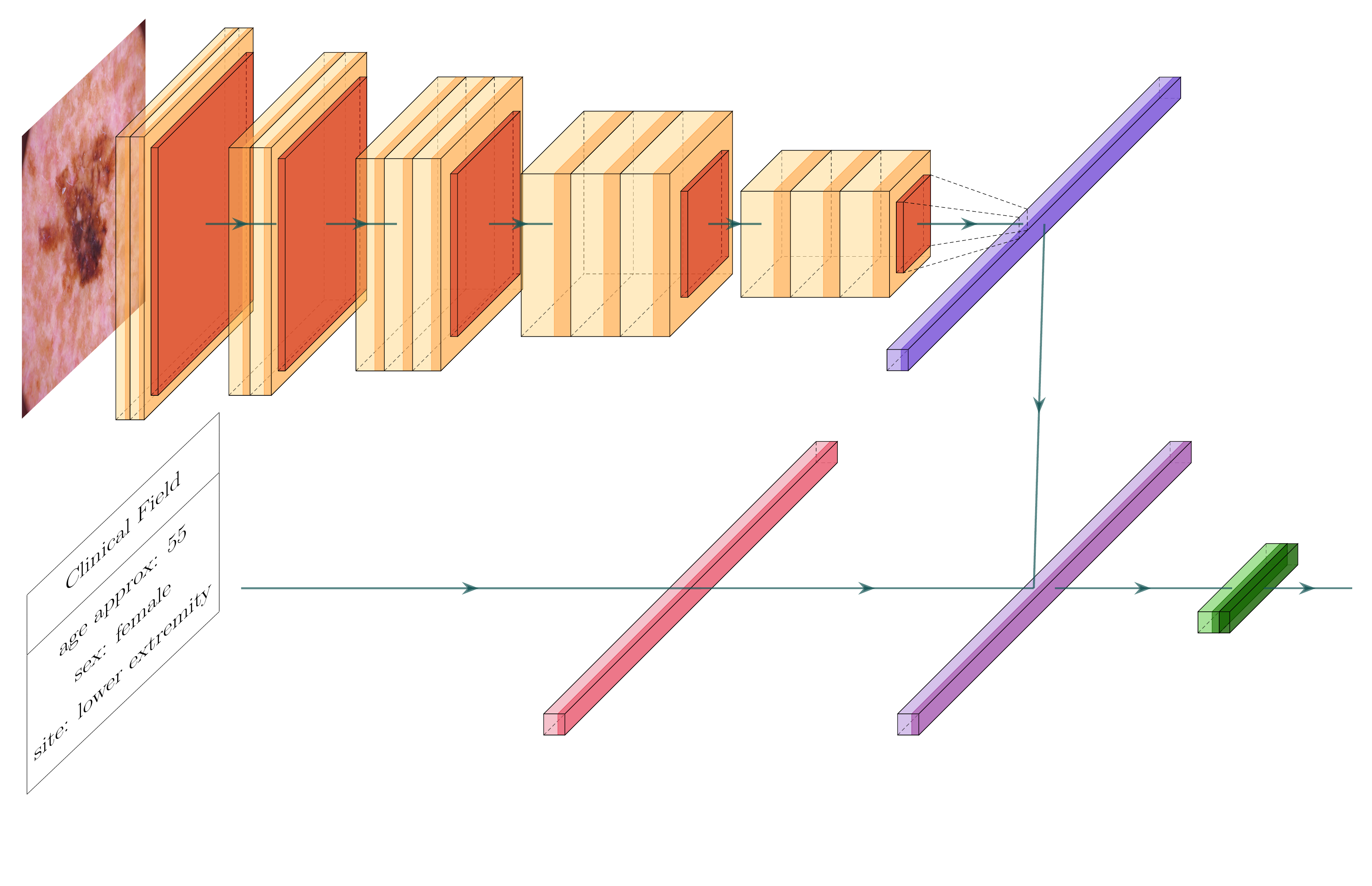}
\caption{Combination of imaging and non-imaging data in deep networks. A series of convolution and pooling operations (orange) yield a lower dimensional feature vector (blue) for image data. The non-imaging data are encoded numerically by mapping to ASCII decimal \cite{Thomas2019CMR} providing a metadata feature vector (red). The imaging and non-imaging feature vectors are concatenated (purple) and used as input for a softmax classifier (green). 
}
\label{fig:network}
\end{figure}

The metadata fields are integrated with the image data by concatenating the image features obtained by the CNN at its deepest layer prior to classification, $\mathcal{F}(X)$ (blue vector in Fig.~\ref{fig:network}), with the encoded metadata inputs, $\mathcal{G}(M)$ (red vector in Fig.~\ref{fig:network}).

\begin{equation}
\label{eq:featureFusion}
\begin{split}
\mathcal{H} &= 
\begin{pmatrix} 
\mathcal{F}(X) \quad \mathcal{G}(M) \\
\end{pmatrix} 
\\
&= 
\begin{pmatrix} 
\mathcal{F}_{1,1} & \cdots & \mathcal{F}_{1,d_K} & \mathcal{G}_{1,1} & \cdots & \mathcal{G}_{1,d_{K'}}\\ 
\vdots & \ddots & \vdots & \vdots & \ddots & \vdots \\ 
\mathcal{F}_{N,1} & \cdots & \mathcal{F}_{N,d_K} & \mathcal{G}_{N,1} & \cdots & \mathcal{G}_{N,d_{K'}}
\end{pmatrix} 
\end{split}~,
\end{equation}
where $N$ is the number of images, $d_K$ is the dimensionality of the output of the neural network, $\mathcal{F}(X)$, and $d_{K'}$ is the dimensionality of the converted metadata, $\mathcal{G}(M)$.

\subsection{Classification}
In all cases we use a softmax function to build a classification model for $K$ classes,
\begin{equation}
\label{eq:softmax}
\sigma(z)_i = \frac{e^{z_{i}}}{\sum_j^K e^{z_{j}}}~.
\end{equation}
This classification model is trained using gradient descent for a maximum of 2000 epochs or when the gradient falls below 10$^{-6}$.  
In this work we compare the performance of the transfer learning based classification of the ISIC image data, to the performance of transfer learning when images are combined with their associated metadata. 
In the former case, we extract the image features, $\mathcal{F}(X)$, from each network pretrained on ImageNet, which are then passed to the softmax function to classify the images. In the latter case, we combine $\mathcal{F}(X)$ and $\mathcal{G}(M)$ as in Eq.~(\ref{eq:featureFusion}), and pass $\mathcal{H}$ to the softmax classifier. In all experiments the data are split into 70:30 training:testing sets that are fixed for all networks for comparability of results. 

We evaluate the performance of our classification models via several metrics. Specifically we evaluate the accuracy, specificity, sensitivity, precision, F-measure, informedness, markendness and Matthews correlation coefficient (MCC). The definitions of these are taken from \cite{Thomas2017} and omitted here for brevity.

%% file: results_short.tex
\section{Results}

\begin{figure}[t!]
\centering
\includegraphics[trim=10em 18em 9em 17em, clip=true,width=0.49\textwidth]{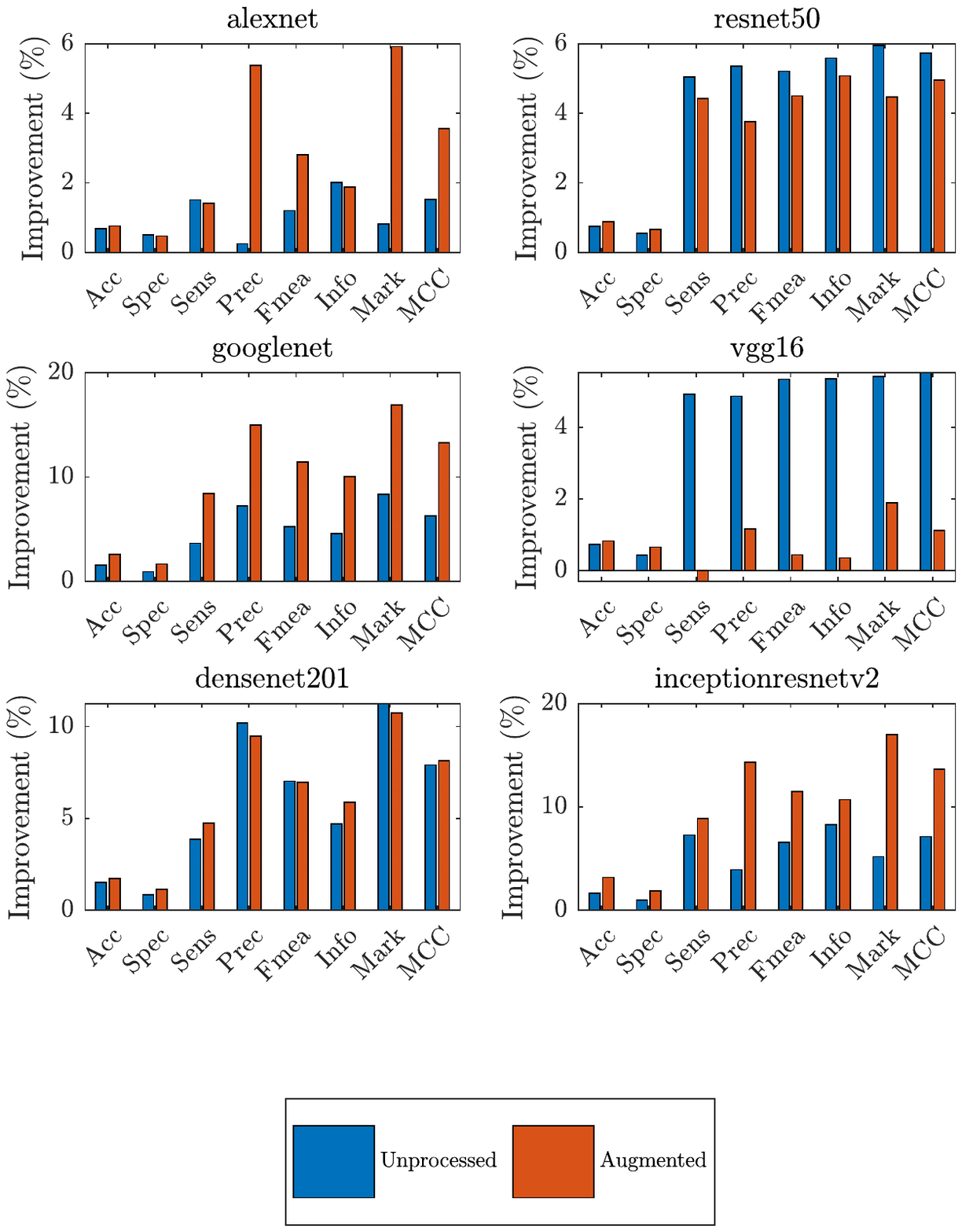}
\caption{Improvement in macro average performance of transfer learning in deep neural networks when using image metadata. Values are the difference in performance scores with positive values demonstrating improved performance when using metadata with image features. For example scores of 70$\%$ (image only) and 80$\%$ (combined image and metadata) would be plotted as 10$\%$. }
\label{fig:MacroAverage}
\end{figure}

We report the macro average (mean class) performance in order to concisely summarise the findings of our experiments. In all of our experiments, we find that combining metadata with the image features improves classification performance for all networks compared to classifying using only image features. This enhancement is observed in all metrics indicating this may be a general characteristic of deep networks. This is clearly illustrated in Fig.~\ref{fig:MacroAverage} where positive values indicate an improvement when including metadata. The only degradation observed was in the sensitivity of a vgg16 network when using data augmentations. However, this decrease is small and this network exhibits relatively low improvements with augmented data compared to the other networks in this work. Improvements in accuracy and specificity are relatively small in all cases, though substantial improvements in the other metrics are seen in all networks. Specifically, googlenet, densenet201 and inceptionresnetv2 show improvements of more than 10 percentage points, meaning a score of 0.7 when using only image data increases to $\geq$ 0.8, a significant improvement.

\begin{figure}[t!]
\centering
\includegraphics[trim=10em 18em 9em 17em, clip=true,width=0.49\textwidth]{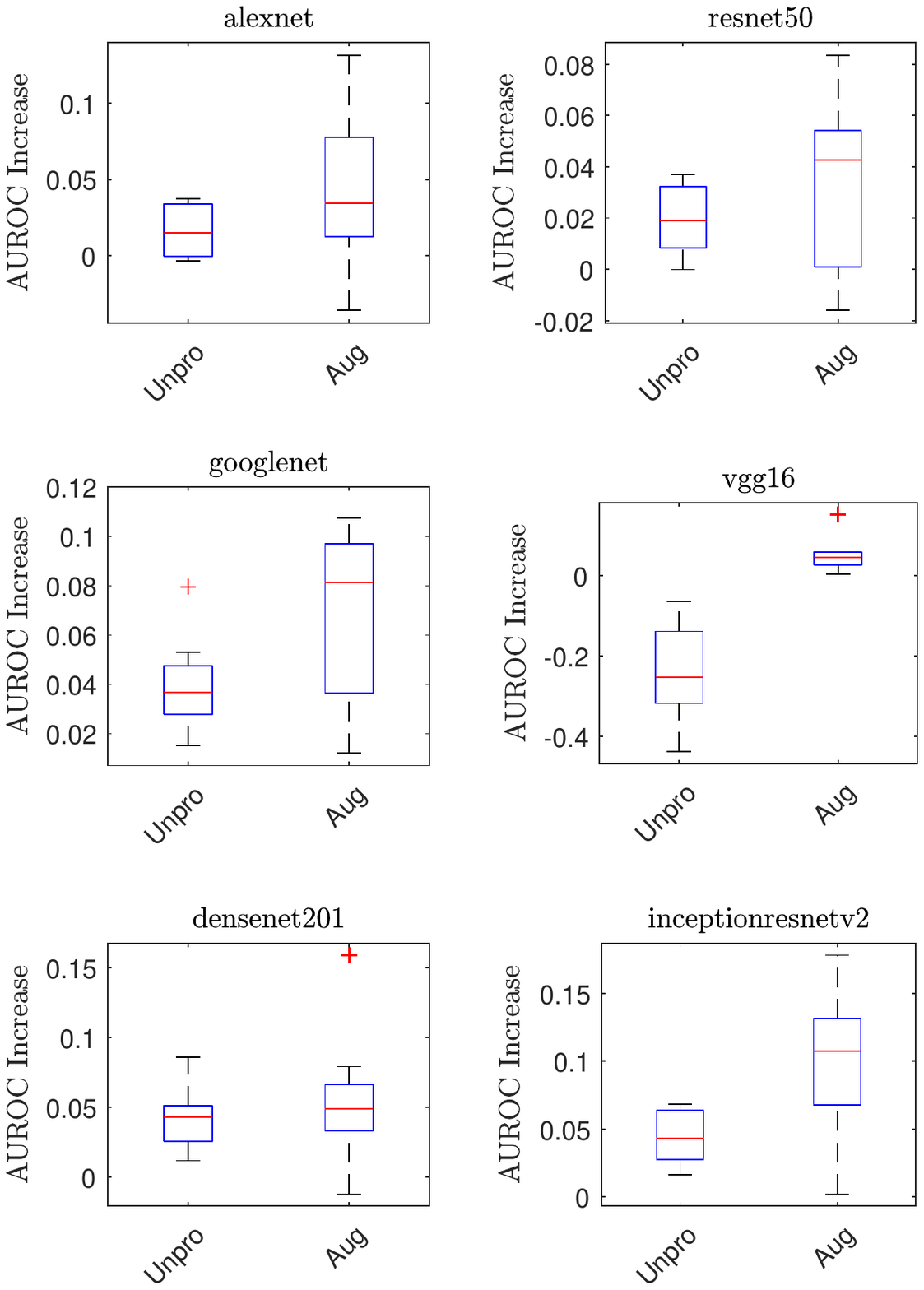}
\caption{Box plots of the class-wise AUROC improvement due to the inclusion of metadata. Values are the difference in AUROC between models that combine image features with metadata and those based only on images features. Positive vales represent enhanced performance and negative values indicate model degradation. Results from both unprocessed (Unpro) and augmented images (Aug) are presented. Note the AUROC ranges from 0 to 1.}
\label{fig:AUC}
\end{figure}

To further evaluate the effects of combining the image metadata with the image features we also consider the area under the receiver operator characteristic curve (AUROC). For each network and experimental set up we perform a class wise ROC analysis, yielding eight receiver operator curves and corresponding AUROCs for each network. We subtract the AUROC for the image data alone from the AUROC when combining the image features and metadata. This class-level measure of improvement or degradation is represented as boxplots presented in Fig.~\ref{fig:AUC}. There is an overall increase in AUROC in all cases except the unprocessed images when using a vgg16 network which shows a considerable degradation. When using augmented images vgg16 shows an enhancement in line with the other networks.

It is worth noting that these improvements come at a negligible cost as seen in Table~\ref{tab:trainingSummary}. The training time for the softmax classifier when using the combined data is comparable to when using the image features alone. Moreover, this is insignificant compared to the feature extraction time in all networks, smaller by up to two orders of magnitude. The low time cost makes this a practical extension of current methods where metadata are available.

\begin{table}[h]
\renewcommand{\arraystretch}{1.3}
\caption{Network Summary of Runtimes. }
\label{tab:trainingSummary}
\centering
\begin{tabular}{|l|c|c|c|c|}
\hline
Network & $d_K$ & Extraction (s) & $\mathcal{F}$ (s) 
& $\mathcal{H}$ (s) \\
\hline 
alexnet             & 4096 & 217; 238   & 24; 21 & 95; 95 \\
resnet50            & 2048 & 2160; 2174 & 22; 22 & 66; 63 \\
googlenet           & 1024 & 780; 800   & 52; 53 & 52; 50 \\
vgg16               & 4096 & 2365; 2323 & 17; 27 & 90; 92 \\
densenet201        & 1920 & 6435; 6362 & 65; 69 & 67; 63 \\
inceptionresnetv2   & 1536 & 5750; 5728 & 34; 38 & 66; 64\\
 \hline 
\end{tabular}
{Extraction is the time to obtain the $d_K$ dimensional features from the network processing over 48 CPU cores. Training times refer to time to train the softmax classifier based on either input features from $\mathcal{F}$ or $\mathcal{H}$. Times for the unprocessed (left) and augmented (right) data are provided respectively for each case.}
\end{table}

%% file: conclusion.tex
\section{Conclusion}

Adding metadata to image features enhances classification overall. These improvements are noted in six different deep convolutional neural networks, as assessed by several performance metrics. Moderate to large enhancements are observed in all networks, with degradation only noted in a vgg16 architecture. Our results indicate that this may be a general property in classification of images with deep neural networks, though more work is required. These improvements come at a negligible additional cost in computation time, and therefore are a practical method for other applications.